\documentclass[10pt,twocolumn,letterpaper]{article}

\usepackage{iccv}
\usepackage{times}
\usepackage{epsfig}
\usepackage{graphicx}
\usepackage{amsmath}
\usepackage{amssymb}

\usepackage{multirow}
\usepackage{multicol}
\usepackage{booktabs}
\usepackage{marvosym}
\usepackage{ifsym}
\usepackage{pifont}
\usepackage{url}

\usepackage[pagebackref=true,breaklinks=true,colorlinks,bookmarks=false]{hyperref}

\usepackage[capitalize]{cleveref}
\crefname{section}{Sec.}{Secs.}
\Crefname{section}{Section}{Sections}
\Crefname{table}{Table}{Tables}
\crefname{table}{Tab.}{Tabs.}

\iccvfinalcopy 

\begin{document}

\title{Real-Aug: Realistic Scene Synthesis for LiDAR Augmentation in 3D Object Detection}

\author{Jinglin Zhan\textsuperscript{1}, Tiejun Liu\textsuperscript{1}, Rengang Li\textsuperscript{1}, Jingwei Zhang\textsuperscript{1}, Zhaoxiang Zhang\textsuperscript{3}, Yuntao Chen\textsuperscript{2}\thanks{Corresponding author: chenyuntao08@gmail.com}\\
\\
\fontsize{11pt}{\baselineskip}\selectfont$^1\text{Inspur Electronic Information Industry Co.,Ltd.}$\\
\fontsize{11pt}{\baselineskip}\selectfont$^2\text{Centre for Artificial Intelligence and Robotics, HKISI, CAS}$\\
\fontsize{11pt}{\baselineskip}\selectfont$^3\text{Institute of Automation, CAS}$\\
}

\maketitle

\begin{abstract}
    Data and model are the undoubtable two supporting pillars for LiDAR object detection.However, data-centric works have fallen far behind compared with the ever-growing list of fancy new models. In this work, we systematically study the synthesis-based LiDAR data augmentation approach (so-called GT-Aug) which offers maxium controllability over generated data samples. We pinpoint the main shortcoming of existing works is introducing unrealistic LiDAR scan patterns during GT-Aug. In light of this finding, we propose Real-Aug, a synthesis-based augmentation method which prioritizes on generating realistic LiDAR scans. Our method consists a reality-conforming scene composition module which handles the details of the composition and a real-synthesis mixing up training strategy which gradually adapts the data distribution from synthetic data to the real one. To verify the effectiveness of our methods, we conduct extensive ablation studies and validate the proposed Real-Aug on a wide combination of detectors and datasets. We achieve a state-of-the-art 0.744 NDS and 0.702 mAP on nuScenes \emph{test} set. The code shall be released soon.
\end{abstract}

\section{Introduction}

The field of autonomous driving has seen a surge of interest in LiDAR 3D object detection due to its ability to overcome the limitations of image-based methods and improve overall system reliability. 
There has been an ever-growing list of novel point cloud feature extractors~\cite{VoxelNet,PointPillar,SECOND,RangeDet,FocalSparseConv,LargeKernel3D,PVRCNN2} and new detection paradigms~\cite{PointPillar,PVRCNN2,CenterPoint,VoxelTransformer} in this area.
In the meantime, data-centric works in this field fall far behind model-centric ones, though data and model are widely recognized as two fundamental components in perception tasks.
The quantity and quality of point cloud data play key roles in achieving a performant detector and data augmentation has always been an integral part of this. 
However, existing works of LiDAR data augmentation either focus more on data under special weather condition~\cite{FogSim, SnowfallSim} or fall short at verifying their effectiveness~\cite{Lidarsim,LidarAug,ShapeAwareAug,StructureAwareAug} on large-scale real-world datasets~\cite{nuScenes,Waymo}.
In this work, we systematically study the synthesis-based approach for LiDAR data augmentation, which indicates the produce of placing a set of object point clouds into scene point clouds~\cite{SECOND,LidarAug}.
Compared with scan-based(e.g. flip, scale, rotate) and object-based LiDAR data augmentation~\cite{PPBA}, synthesis-based ones generate diverse LiDAR scans and offer fine-grain controllability over the synthesized scenes like over-sampling objects from rare classes.

\begin{figure}[t]
\begin{center}
   \includegraphics[width=1.0\linewidth]{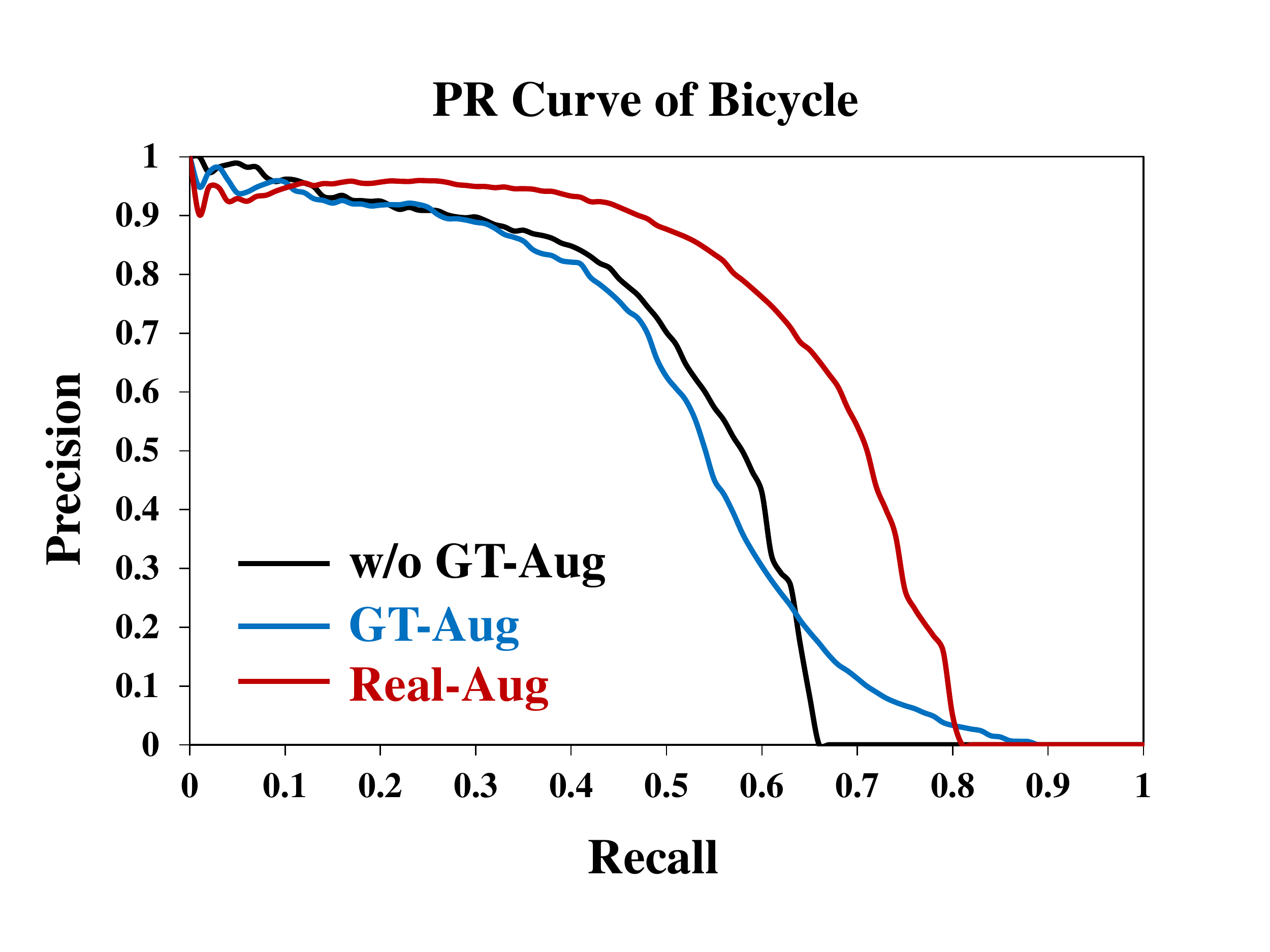}
\end{center}
   \caption{Pr-curve of the bicycle at different scene synthesis steps.}
\label{figure: pr curve of bicycle}
\end{figure}

However, simply applying the vanilla synthesis-based LiDAR data augmentation(so-called GT-Aug) does not lead to satisfactory results on modern large-scale datasets like nuScenes and Waymo~\cite{db_sampler_issue_in_centerpoint}.
To explain the above phenomenon, we take the bicycle class in nuScenes as an example and plot its PR-Curve for a CenterPoint~\cite{CenterPoint} detector trained with and without GT-Aug. 
Fig.~\ref{figure: pr curve of bicycle} shows that after applying GT-Aug, the detector is able to recall more objects at the cost of generating more false positives.
The PR-curve clearly reveals a downside of vanilla GT-Aug - introducing non-existing LiDAR scans pattern into the original dataset. 

Real-Aug, which prioritizes on the realisticness of newly synthesized scenes, is proposed in this paper to overcome the limitation of vanilla synthesis-based LiDAR augmentation methods.
It consists a reality-conforming scene composition module for handling intricate technical details throughout scene composition and a real-synthesis mixing up training strategy which gradually aligns the distribution from synthetic data to the real one. 
The effectiveness of Real-Aug are validated across multiple LiDAR object detection datasets for different detectors.
We achieve a 4.7\% 3D mAP improvement on KITTI 3D object detection benchmarks (2.1\%, 4.0\%, 8.1\% for car, pedestrian and cyclist respectively) for a baseline SECOND~\cite{SECOND} detector.
Notably, we achieve a 74.4\% NDS and a 70.2\% mAP on the $test$ set of nuScenes 3D object detection benchmark\footnote{Rank 1st among all LiDAR-only object detection methods by the time of submission}. There is a significant 6.1\% NDS and a 8.7\% mAP improvement over our baseline CenterPoint~\cite{CenterPoint} detector solely through data augmentation. 

Our contributions can be summarized as follows:

(1) We reveal the realisticness issue of vanilla synthesis-based LiDAR data augmentation.

(2) We present a well-designed Real-Aug scheme, which features a reality-conforming scene composition module and a real-synthesis mixing up strategy.

(3) We highlight the effectiveness of Real-Aug by applying it cross a wide range of detectors for multiple datasets and achieve state-of-the-art results. 

(4) We validate technical advantages of Real-Aug, including its robustness to hyper parameter choices and the improvement of data utilization.  
    
\section{Related Work}

\subsection{Non-synthesis Data Augmentation Methods}

Data augmentaion methods are widely applied to artificially diversify the dataset and help promote the detectors' capacity. Commonly-used strategies include random flip, rotation, scale, and translation at both scene- and instance- level. Some physically valid simulation methods were deployed to deal with the detection challenges under foggy or snowy weather~\cite{FogSim,SnowfallSim}. PointPainting augmented point clouds with image semantics. It appended the predicted class score from image semantic segmentation network to each point~\cite{PointPainting}. Inspired by PointPainting, PointAugmenting decorated point clouds with corresponding point-wise CNN features extracted from 2D image detectors~\cite{PointAugmenting}.

\subsection{Synthesis-based Data Augmentation Methods}

Mix3D devised a "mixing" technique to create new scenes by combining two augmented ones while ensuring sufficient overlap~\cite{Mix3D}. A sensor-centric approach was applied for maintaining the data structure of synthesized scenes consistent with the lidar sensor capacities~\cite{StructureAwareAug}. 
One of the most popular synthesis-based data augmentation methods, Ground-Truth augmentation (GT-Aug), was presented by Yan et al.~\cite{SECOND} in 2018 and applied in multiple LiDAR detection tasks~\cite{PointPillar, CenterPoint, CBGS, LidarMultiNet, LargeKernel3D, AFDetV2, MDRNet, TransFusion}. On top of GT-Aug, many techniques were proposed for diversifying the ground-truth database.
Part-aware and shape-aware gt sampling divided objects into partitions and stochastically applied augmentation methods to each local region~\cite{PartAwareAug, ShapeAwareAug}. Pattern-aware gt sampling downsampled the points of objects to create a new one with farther distance~\cite{PatternAwareAug}. 
PointMixup utilized an interpolation method to compose new objects~\cite{PointMixup}. 
PointCutMix replaced part of the sample with shape-preserved subsets from another one~\cite{PointCutMix}. Fang et al. proposed a rendering-based method for inserting visual objects simulated by CAD into the real background~\cite{LidarAug}.

Placing instances at semantically plausible positions was proved to be essential to guarantee the improved performance for 2D object detectors~\cite{ImageObjectPlace1,ImageObjectPlace2,ImageObjectPlace3}. In LiDAR-based 3D object detection, collision problem is commonly seen as a physical placement issue in GT-Aug. Yan et al. performed a collision test after ground-truth sampling and removed any sampled objects that collided with others~\cite{SECOND}. Competition strategy, which remains the points closer to the sensor, was employed to generate a more physical synthesized scene~\cite{StructureAwareAug}. 
LiDAR-Aug leveraged a "ValidMap" to generate poses for achieving more reasonable obstacle placements~\cite{LidarAug}. It divided point clouds into pillars and filtered out valid pillars according to the height distribution. Although some researchers have noticed the placement issue in GT-Aug, the systematical studies about unrealistic LiDAR scan patterns in synthesized scenes is still woefully insufficient. Particularly, the deviations of data distribution from synthetic data to the real one are rarely discussed. 
As a result, existing synthesis-based LiDAR augmentations only achieved limited success, especially in large-scale datasets like nuScenes and Waymo. 

\section{Realistic Scene Synthesis}
\label{sec:realistic scene synthesis}
Our method mainly consists of a reality-conforming scene composition module and a real-synthesis mixing up training strategy.
We introduce a reality-conforming score in Sec.~\ref{sec:reality_conforming_score} to measure the realisticness of synthesized scans.
The details of scene composition is described in Sec.~\ref{sec:scene_composition}. 
We elaborate the training strategy of how to blend synthesized and real LiDAR scans to achieve the optimal performance in Sec.~\ref{sec:real-synthesis mixing up}.

\begin{figure*}[t]
\begin{center}
   \includegraphics[width=1.0\linewidth]{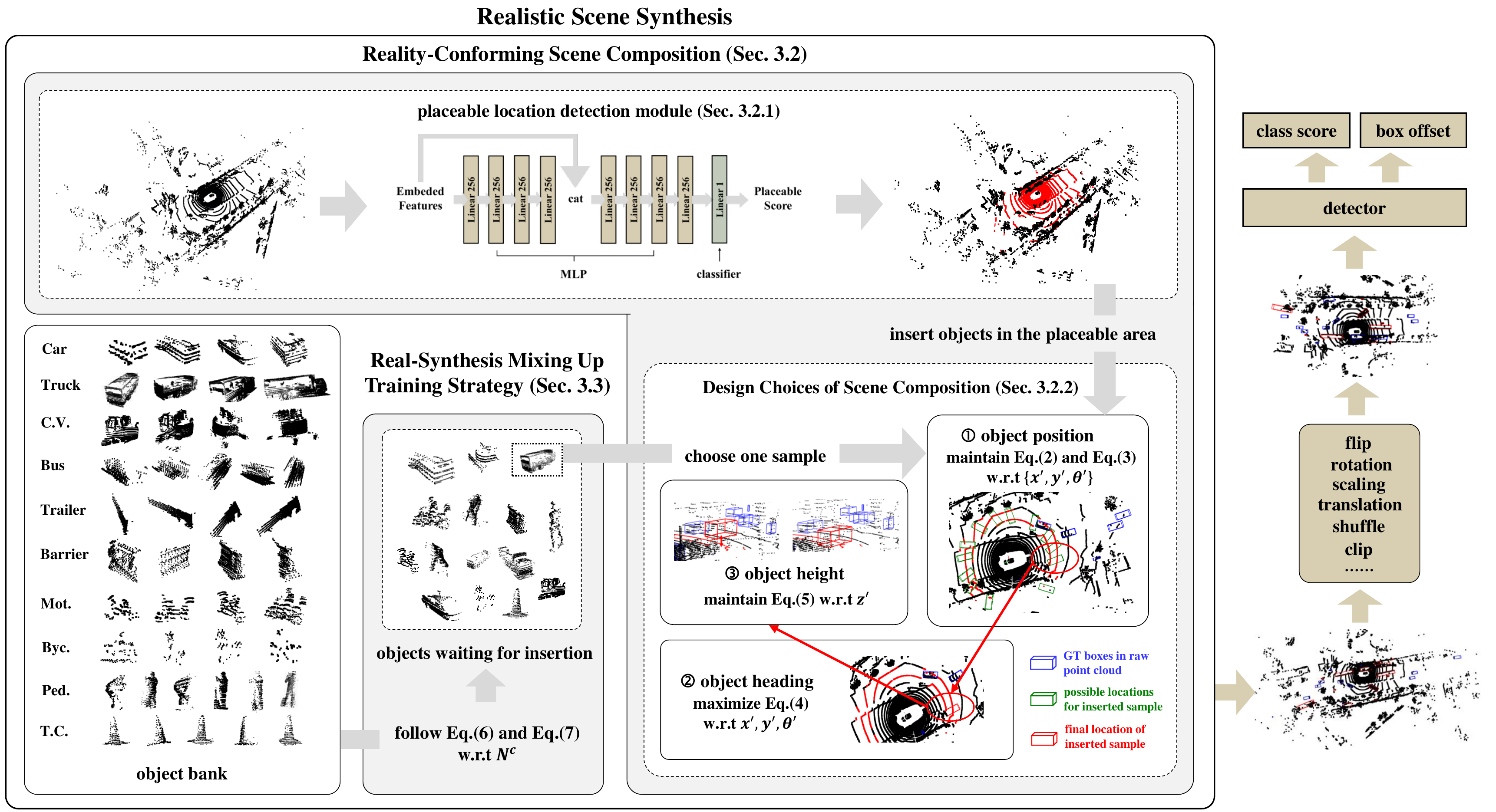}
\end{center}
   \caption{Overview of the proposed realistic scene synthesis for LiDAR augmentation (Real-Aug) in 3D object detection. (Points reflected from placeable area is painted with red. GT boxes in raw point cloud, possible locations for inserted sample and final location of inserted sample are presented by the bounding boxes with blue, green and red lines respectively.)}
\label{figure: main graph}
\end{figure*}
\subsection{Reality-Conforming Score}
\label{sec:reality_conforming_score}
Finding a proper metric to measure the realisticness is at the core of our method.
The most direct approach is measuring how well a model trained on the vanilla \emph{train} set perform on the augmented \emph{val} set.
If the newly synthesized scenes conform to the same data distribution of the train set, the vanilla model could recognize it without any performance degradation.  
Therefore, we define a reality-conforming score $Re$ directly based on metric of the perception task at hand, which could seen as a generalized version of detection agreement score mentioned in LiDARSim~\cite{Lidarsim}.
Specifically, for LiDAR object detection, we define the the reality-conforming score $Re(\text{mAP})$ as the ratio between mAP tested on $val$ set with and without ground-truth augmentation. 

\begin{equation}
    Re(\text{mAP}) = \dfrac{\text{mAP}_{\text{aug}}}{\text{mAP}_{\text{noaug}}}
\end{equation}

The reality-conforming score could also be defined over other metrics, like mIoU for semantic segmentation and PQ for panoptic segmentation.

\subsection{Reality-Conforming Scene Composition}
\label{sec:scene_composition}
We perform our scene synthesis solely via the object-scene composition approach, which could be formulated as sequentially place LiDAR points of one or more objects into a existing scene.
We refer the set of all placeable objects as the \emph{object bank} and all scenes as the \emph{scene bank} accordingly.

The composition approach is widely adopt in both 2D and 3D object detection~\cite{CutMix, ImageCutMixUse, SwinTransformer, SECOND, CenterPoint} but achieves far less success in large scale LiDAR object detection benchmark like nuScenes~\cite{nuScenes} and Waymo ~\cite{Waymo}.
We find the realisticness is the key to the success of this kind of methods and elaborate the technical details affecting the realisticness of synthesized LiDAR scans in the following sections. 
Related ablation studies could be found in Tab.~\ref{table: nusc reality conforming score}

\subsubsection{Placeable Location Detection Module}
\label{sec:drivable areas detector}
It is obvious that not everywhere in a LiDAR scan is a suitable spot for placing an object.
Accurate modeling where each kind of object could appear requires a ton of extra knowledge, we simplify this task by assuming most objects of interest are on the drivable surface.
This simplification is not perfect in every case(e.g. a pedestrian could appear on the sidewalk) but it is a good approximation as objects appeared in the drivable area affect the behavior of ego vehicle most.

We adopt a light-weight coordinate MLP~\cite{Nerf} as our placeability estimator.
The input of the network is the coordinates and reflectivity of LiDAR points $(x,y,z,r)$. 
We use a Fourier~\cite{Nerf} encoder of order $L=10$ to map the points into 64-dim embeddings. 
We use binary cross-entropy loss for training the estimator. 
The supervision could come from either model-based ground estimation method like PatchWorks~\cite{Patchwork, Patchwork++} or manually labelled LiDAR semantic segmentation.

An alternative way would be directly using ground estimator like PatchWorks but we choose the coordinated-MLP apporach for its denoising nature and low latency($<$ 1ms) on modern hardware.

\subsubsection{Design Choices of Scene Composition}

\noindent\textbf{Object Position.}
Different from the vanilla GT-Aug which always place the object at the same position from where it been take, there are three factors in Real-Aug to consider when choosing physically reasonable position of a sampled object: the distance and observing angle from ego vehicle, as well as the predicted placeability.

Assuming the XOY location and heading of an object in its original scene is denoted as $(x, y, \theta)$ and its location and heading in the synthesized scene as $(x', y', \theta')$, the distance constraint could be specified as,
\begin{equation}
\label{eq:distance constraint}
    x^2 + y^2 = x'^2 + y'^2 \pm \Delta
\end{equation}
and observing angle constraint as
\begin{equation}
\label{eq:observing angle constraint}
    \theta + \arctan\dfrac{y}{x} = \theta' + \arctan\dfrac{y'}{x'}.
\end{equation}

Here $\Delta$ is the error tolerance threshold as finding an exact match for distance constraint is almost impossible for a limited number of points in a single scan. 
By default $\Delta$ is set to the half length of object $L/2$. 

The distance constraint ensures the realistic point density and the observing angle constraint guarantees the a realistic scan pattern.
Finally for the placeable constraint, we simply reject all locations $(x', y')$ with predicted placeability $< 0.5$ to avoid placing objects into unfavorable locations.

\noindent\textbf{Object Heading.}
Taking a closer look at Eq.~\ref{eq:distance constraint} and Eq.~\ref{eq:observing angle constraint} we can find that the heading angle $\theta'$ of our object is a free variable, which opens up the possibility of choosing a natural heading angle for the object instead of placing it into the scene with some wired random headings.
We select the heading angle $\theta'$ of our object to conform the heading distribution of objects from the same category in the scene$\{\theta_c\}$ via measuring the cosine similarity between headings,
\begin{equation}
    \theta' = {\arg\max}_\theta \sum_c \cos(\theta - \theta_c)
\end{equation}
If there is no object in the scene with the same category, we simply choose the most frequent heading for our object.

\noindent\textbf{Object Height.}
Using the original height of an object could make it fly over or under the new scene's ground plane. We mitigate this by setting its bottom height in the new scene as the mean height of all ground points $\{z_{g}\}$ enclosed by the bounding box.

\begin{equation}
    z'-H/2 = {\rm avg}(z_g)
\end{equation}
\noindent\textbf{Collision Avoidance.}
In order to avoid collision, we use the same strategy in GT-Aug\cite{SECOND} and remove placed object if it overlaps with the existing ones. 

\subsection{Real-Synthesis Mixing Up Training Strategy}
\label{sec:real-synthesis mixing up}

In this section, we elaborate a real-synthesis mixing up training strategy for gradually adapting the detector from synthetic data distribution to the real one.
We introduce the real scene-category relation and scene-crowdedness relation alignments in Sec.~\ref{sec:align to real scene-category relation} and Sec.\ref{sec:align to real scene-crowd relation} to fulfill the full potential of Real-Aug.

\subsubsection{Align to Realistic Scene-Category Relation}
\label{sec:align to real scene-category relation}

\setlength{\tabcolsep}{5pt}
\setlength{\doublerulesep}{2\arrayrulewidth}
\renewcommand{\arraystretch}{1.0}
\begin{table}[ht]
    \begin{center}
    \begin{tabular}{c|c|c|c}
        \toprule
        \multirow{2}{*}{class} & scene-0184 & scene-0234 & scene-0399 \\
         & (residential) & (downtown) & (expressway) \\
        \hline
        Car & 54 & 714 & 1155 \\
        Truck & 0 & 217 & 115 \\
        C.V. & 0 & 0 & 0 \\
        Bus & 0 & 59 & 41 \\
        Trailer & 0 & 80 & 0 \\
        Barrier & 0 & 95 & 0 \\
        Mot. & 0 & 0 & 13 \\
        Byc. & 13 & 0 & 0 \\
        Ped. & 0 & 1404 & 0 \\
        T.C. & 0 & 83 & 59 \\
        \bottomrule
    \end{tabular}
    \end{center}
    \caption{The category distributions of objects at scene-0184, scene-0234, scene-0399.}
    \label{table: scene-class relation}
\end{table}

Existing large-scale autonomous driving datasets~\cite{Argoverse,Waymo,nuScenes,kitti} make great efforts at ensuring the diversity of video clips and generally contains scenarios from a wide range of weather, lighting and road conditions.
However, the rich diversity of scenarios poses challenge for scene-synthesis.
For instance, putting a bike rider on a closed cross-state highway or synthesizing a man holding an umbrella into a sunny afternoon country road both make the synthesized scenes highly unrealistic.
So maintaining a reasonable scene-category relation is of great importance in our work.

We summarize the scene-category relation for three 20s video clips from nuScenes in Tab.\ref{table: scene-class relation}. 
It reveals strong connection between category distribution of objects and their surrounding environments. 
Previous approaches like GT-Aug fail to realize the scene-category relation in driving scenarios, leading to detectors which hallucinate non-existing false positives as shown in Fig.~\ref{figure: pr curve of bicycle}.

In order to both enjoy the enhanced feature learning via scene augmentation and respect the original scene-category relation, we propose a mix up training strategy.
The plain strategy is to place a preset number of objects for each category into the scene for each LiDAR scan. 
Using $c$ to denote the object category, the number of inserted objects can be represented as $N_\text{plain}^{c}$. 
The plain strategy totally ignores the scene-category relation but generates diverse scans which is good for feature learning.
Another strategy is to strictly respect the scene-category relation by inserting objects only from the existing categories in the scan. We use $N_\text{exist}^{c}$ to denote the number of objects inserted by this strategy, where $N_\text{exist}^{c} = 0$ for categories not exist on this scan.
We use a hyper-parameter $\alpha \in [0, 1]$ to balance above two strategies and obtain our final strategy $N^c$. 
We align the data distribution to the real scene-category relation by gradually annealing $\alpha$ from 1 to 0 towards the end of training. 

\begin{equation}
\label{eq:align real scene-category relation}
    N^c = N_\text{plain}^c \times \alpha + N_\text{exist}^c \times (1-\alpha)
\end{equation}

\subsubsection{Align to Realistic Scene-Crowdedness Relation}
\label{sec:align to real scene-crowd relation}

\setlength{\tabcolsep}{3pt}
\setlength{\doublerulesep}{2\arrayrulewidth}
\renewcommand{\arraystretch}{1.0}
\begin{table}[ht]
    \begin{center}
    \begin{tabular}{c|c|c|c}
        \toprule
        class & fg/bg(w/o GT-Aug) & fg/bg(w/ GT-Aug) & Ratio \\ 
        \hline
        Car & 0.791\% & 0.799\% & 1.01 \\
        Truck & 0.193\% & 0.306\% & 1.59 \\
        C.V. & 0.042\% & 0.621\% & 14.79 \\
        Bus & 0.060\% & 0.394\% & 6.57 \\
        Trailer & 0.095\% & 0.634\% & 6.67 \\
        Barrier & 0.246\% & 0.336\% & 1.39 \\
        Mot. & 0.021\% & 0.378\% & 18.00 \\
        Byc. & 0.020\% & 0.384\% & 19.20 \\
        Ped. & 0.344\% & 0.379\% & 1.10 \\
        T.C. & 0.143\% & 0.231\% & 1.62 \\
        \bottomrule
    \end{tabular}
    \end{center}
    \caption{Comparing the ratios of foreground voxels to background voxels with and without GT-Aug. The resolution used here for voxelization is [0.075m,0.075m,0.2m] and number of voxels are counted on a feature map of a total stride of 8.}
    \label{table: fg and bg ratio}
\end{table}

While our composition-based augmentation greatly facilitate the feature leanring for LiDAR object detector, it inevitably distorting the crowdedness of the real scenes as we only inserting objects into scenes.

Tab.~\ref{table: fg and bg ratio} demonstrates the increase of foreground voxels on the feature of a CenterPoint detector when applying GT-Aug for LiDAR augmentation. 
Among 10 categories defined in nuScenes detection task, fg/bg of bicycle, motorcycle and construction vehicle  rank the top three with increased times of 19.2, 18.0, 14.8. 
The significant increase of foreground voxels encourages detectors to make more predictions than what there really are.
To deal with this, we introduce another hyper-parameter $\beta$ and also use a gradually annealing strategy to decrease its value for aligning the real scene-crowdedness relation. 

\begin{equation}
    N^c = (N_\text{plain}^c \times \alpha + N_\text{exist}^c \times (1-\alpha) )\times \beta
\end{equation}

\section{Experiments}

For verifying the effectiveness and generality of Real-Aug, we conduct extensive experiments on nuScenes and KITTI. 
Some brief descriptions of datasets are summarized in Sec.~\ref{sec:datasets and metrics}. 
The implement details are shown in Sec~\ref{sec:implement details}. 
We elaborate the evalutation results on the $test$ set of nuScenes and KITTI in Sec.~\ref{sec:evaluation on nuScenes and KITTI test set}.
Exhaustive ablations are performed and discussed in Sec.~\ref{sec:abliations and anaylsis}.

\subsection{Datasets and Metrics}
\label{sec:datasets and metrics}

\textbf{nuScenes Dataset\cite{nuScenes}}. nuScenes dataset is a large-scale dataset designed for accelerating researches on multiple tasks in autonomous driving scenarios. It comprises over 1,000 scenes, which are divided into 700 scenes for training, 150 scenes for validation and 150 scenes for testing. The full dataset consists 390k 360-degree LiDAR sweeps, which are collected by Velodyne HDL-32E with 20Hz capture frequency. The nuScenes detection task requires detecting 10 object classes with 3D bounding boxes, attributes, and velocities. 
For evaluation, the official detection metrics, including NuScenes Detection Score (NDS) and mean Average Precision (mAP), are used.

\textbf{KITTI Dataset\cite{kitti}}. KITTI dataset is a widely used benchmark dataset for 3D object detection. It contains 7481 frames for training and 7518 frames for testing. As Ref \cite{PointPillar, SECOND}, the training frames are further divided into \emph{train} set with 3712 frames and \emph{val} set with 3769 frames. The KITTI detection task requires detecting 3 object classes with 3 difficulty levels (Easy, Moderate, Hard). Detectors are evaluated by 3D Average Precision $AP_{3D}$, which is calculated with recall 40 positions (R40).

\setlength{\tabcolsep}{3pt}
\setlength{\doublerulesep}{2\arrayrulewidth}
\renewcommand{\arraystretch}{1.1}
\begin{table*}[ht]
    \begin{center}
    \begin{tabular}{c|cc|cccccccccc}
        \toprule
        Methods & NDS & mAP & Car & Truck & C.V. & Bus & Trailer & Barrier & Mot. & Byc. & Ped & T.C. \\ 
        \hline
        CBGS \space\cite{CBGS}  & 0.633 & 0.528 & 0.811 & 0.485 & 0.105 & 0.549 & 0.429 & 0.657 & 0.515 & 0.223 & 0.801 & 0.709 \\
        PillarNet-34 \dag\space\cite{PillarNet} & 0.714 & 0.660 & 0.876 & 0.575 & 0.279 & 0.636 & 0.631 & 0.772 & 0.701 & 0.423 & 0.873 & 0.833 \\
        LidarMultiNet \space\cite{LidarMultiNet} & 0.716 & 0.670 & 0.869 & 0.574 & 0.315 & 0.647 & 0.610 & 0.735 & 0.753 & 0.476 & 0.872 & 0.851 \\
        Transfusion\_L \dag\space\cite{TransFusion} & 0.702 & 0.655 & 0.862 & 0.567 & 0.282 & 0.663 & 0.588 & 0.782 & 0.683 & 0.442 & 0.861 & 0.820 \\
        LargeKernel3D\_L\space\cite{LargeKernel3D} & 0.705 & 0.653 & 0.859 & 0.553 & 0.268 & 0.662 & 0.602 & 0.743 & 0.725 & 0.466 & 0.856 & 0.800 \\
        LargeKernel3D\_L \dag\space\cite{LargeKernel3D} & 0.728 & 0.688 & 0.873 & 0.591 & 0.302 & 0.685 & 0.656 & 0.750 & 0.778 & 0.535 & 0.883 & 0.824 \\
        AFDetV2 \space\cite{AFDetV2} & 0.685 & 0.624 & 0.863 & 0.542 & 0.267 & 0.625 & 0.589 & 0.710 & 0.638 & 0.343 & 0.858 & 0.801 \\
        MDRNet\_L \space\cite{MDRNet} & 0.705 & 0.652 & 0.865 & 0.545 & 0.257 & 0.638 & 0.589 & 0.748 & 0.731 & 0.452 & 0.866 & 0.829 \\
        MDRNet\_L \dag\space\cite{MDRNet} & 0.720 & 0.672 & 0.873 & 0.577 & 0.283 & 0.665 & 0.622 & 0.752 & 0.744 & 0.485 & 0.876 & 0.843 \\
        \hline
        CenterPoint\space\cite{CenterPoint}  & 0.655 & 0.580 & 0.846 & 0.510 & 0.175 & 0.602 & 0.532 & 0.709 & 0.537 & 0.287 & 0.834 & 0.767  \\
        \textbf{CenterPoint+Real-Aug} & \textbf{0.709} & \textbf{0.658} & \textbf{0.852} & \textbf{0.546} & \textbf{0.313} & \textbf{0.652} & \textbf{0.600} & \textbf{0.770} & \textbf{0.726} & \textbf{0.464} & \textbf{0.857} & \textbf{0.800} \\ 
        CenterPoint \dag\space\cite{CenterPoint} & 0.673 & 0.603 & 0.852 & 0.535 & 0.200 & 0.636 & 0.560 &	0.711 & 0.595 & 0.307 & 0.846 & 0.784 \\
        \textbf{CenterPoint+Real-Aug} \dag & \textbf{0.734} & \textbf{0.690} & \textbf{0.858} & \textbf{0.582} & \textbf{0.349} & \textbf{0.673} & \textbf{0.639} & \textbf{0.787} & \textbf{0.784} & \textbf{0.523} & \textbf{0.881} & \textbf{0.811} \\
        \textbf{SparseFishNet3D+Real-Aug} \dag & \textbf{0.744} & \textbf{0.702} & \textbf{0.868} & \textbf{0.593} & \textbf{0.355} & \textbf{0.701} & \textbf{0.656} & \textbf{0.776} & \textbf{0.783} & \textbf{0.551} & \textbf{0.890} & \textbf{0.845} \\ 
        \bottomrule
    \end{tabular}
    \end{center}
    \caption{Comparison with state-of-the-art methods on $test$ sets of nuScenes detection benchmarks.(\dag: test-time augmentation.)}
    \label{table: nusc test cmp}
\end{table*}

\setlength{\tabcolsep}{3pt}
\setlength{\doublerulesep}{2\arrayrulewidth}
\renewcommand{\arraystretch}{1.0}
\begin{table*}[ht]
    \begin{center}
    \begin{tabular}{c|ccc|ccc|ccc|c}
        \toprule
        Method & \multicolumn{3}{c|}{Car} & \multicolumn{3}{c|}{Pedestrian} & \multicolumn{3}{c|}{Cyclist} & mAP \\ 
        \hline
        SECOND\space\cite{kitti_second_7862} & 85.3\% & 76.6\% & 71.8\% & 43.0\% & 35.9\% & 33.6\% & 71.1\% & 55.6\% & 49.8\%  & 58.1\% \\
        SECOND+Real-Aug & 86.8\% & 78.4\% & 74.7\% & 47.2\% & 40.3\% & 37.2\% & 81.4\% & 63.2\% & 56.2\% & 62.8\% \\
        \bottomrule
    \end{tabular}
    \end{center}
    \caption{$mAP_{3D}$ difference of well-trained SECOND model (using GT-Aug or Real-Aug) on KITTI \emph{test} sets.}
    \label{table: kitti test cmp}
\end{table*}

\subsection{Implement Details}
\label{sec:implement details}

Our implementation of LiDAR-based 3D object detection is based on open-sourced OpenPCDet~\cite{OpenPCDet} and the published code of CenterPoint~\cite{CenterPoint}. For nuScenes, we choose CenterPoint-Voxel, CenterPoint-Pillar, SECOND-Multihead frameworks for analysis. For KITTI, we choose SECOND and PointPillar frameworks for anaylsis. 
Detectors are trained with a batch size of 32 on 8 A100 GPUs. We utilize adam optimizer with one-cycle learning rate policy. 
We use the same data augmentation methods (except GT-Aug) and network designs as prior works~\cite{CenterPoint, CBGS, SECOND, PointPillar}. The total training epochs for nuScenes and KITTI are set as 20 and 80 respectively. We adopt weighted Non-Maxima Suppression (NMS)~\cite{RangeDet} during inference. 
The placeability estimator described in Sec.~\ref{sec:drivable areas detector} is supervised by the ground labels generated from PatchWorks~\cite{Patchwork,Patchwork++}.

\subsection{Evaluation on nuScenes and KITTI \emph{test} set}
\label{sec:evaluation on nuScenes and KITTI test set}

\textbf{nuScenes}. As shown in Tab.~\ref{table: nusc test cmp}, the CenterPoint detector trained with Real-Aug outperforms other state-of-the-art LiDAR-only methods on the nuScenes $test$ set. 
Comparing to the work reported by Yin et al.~\cite{CenterPoint}, Real-Aug promotes the NDS and mAP of CenterPoint from 0.673, 0.603 to 0.734, 0.690. 
Notably, our methods bring significant mAP improvement for bicycle, motorcycle and construction vehicle by 21.6\%, 18.9\%, 14.9\% over the baseline. 
Combining with the SparseFishNet3D backbone which is described in Sec.\ref{sec:backbone optimization}, we achieve 0.744 NDS and 0.702 mAP on nuScenes \emph{test} set. 

\textbf{KITTI}. 
The evaluation metric of KITTI changes from $AP_{3D}$ R11 to $AP_{3D}$ R40. 
For an unbiased comparison, we take the submitted results which are achieved based on the reimplement of OpenPCDet~\cite{kitti_second_7862} as a reference. 
The results shown in Tab.~\ref{table: kitti test cmp} authenticate the effectiveness of Real-Aug in KITTI dataset.
There is an average boost of 4.7\% $AP_{3D}$ for all classes with different difficulties (2.1\%, 4.0\%, 8.1\% for car, pedestrian and cyclist respectively).

\subsection{Ablations And Analysis}
\label{sec:abliations and anaylsis}
Real-Aug is investigated with extensive ablation experiments on \emph{val} set of nuScenes.
In Sec.~\ref{sec:reality-conforming score}, we discuss the realisticness of synthesized scenes.
The advantages of Real-Aug, including its effectiveness, robustness and its role in promoting the data utilization, are analyzed from Sec.~\ref{sec:effectiveness of real-aug on nuscenes val set} to Sec.~\ref{sec:choice of training strategy}. The optimized backbone, which is called SparseFishNet3D, is described in Sec.~\ref{sec:backbone optimization} for achieving better detection performance.

\subsubsection{Reality-Conforming Score}
\label{sec:reality-conforming score}

\setlength{\tabcolsep}{1pt}
\setlength{\doublerulesep}{2\arrayrulewidth}
\renewcommand{\arraystretch}{1.0}
\begin{table}[ht]
    \begin{center}
    \begin{tabular}{c|cccc|cc}
        \toprule
         Methods & Position & Heading & Height & Category & $Re(\text{mAP})$ \\
         \hline
         w/o GT-Aug &  &  &  &  & 1.000\\
         GT-Aug &  &  &  &  & 0.744\\
         Real-Aug & \ding{51} & \ding{51} & \ding{51} & \ding{51} & 0.933\\
         \hline
         Real-Aug &  & \ding{51} & \ding{51} & \ding{51} & 0.870\\ 
         Real-Aug & \ding{51} &  & \ding{51} & \ding{51} & 0.880\\
         Real-Aug & \ding{51} & \ding{51} &  & \ding{51} & 0.906\\
         Real-Aug & \ding{51} & \ding{51} & \ding{51} &  & 0.795\\
        \bottomrule
    \end{tabular}
    \end{center}
    \caption{$Re(\text{mAP})$ of synthesized scenes generated by different methods.}
    \label{table: nusc reality conforming score}
\end{table}

\setlength{\tabcolsep}{3pt}
\setlength{\doublerulesep}{2\arrayrulewidth}
\renewcommand{\arraystretch}{1.1}
\begin{table*}[ht]
    \begin{center}
    \begin{tabular}{c|c|cc|cccccccccc}
        \toprule
         & Methods & NDS & mAP & Car & Truck & C.V. & Bus & Trailer & Barrier & Mot. & Byc. & Ped & T.C. \\ 
        \hline
        0 & GT-Aug & 0.666 & 0.595 & 0.849 & 0.586 & 0.184 & 0.696 & 0.389 & 0.687 & 0.587 & 0.427 & 0.850 & 0.696 \\
        \hline
        1 & w/o GT-Aug & 0.659 & 0.594 & 0.837 & 0.571 & 0.170 & 0.677 & 0.372 & 0.678 & 0.648 & 0.449 & 0.847 & 0.696 \\
        2 & + Real Composition & 0.678 & 0.611 & 0.851 & 0.593 & 0.206 & 0.722 & 0.425 & 0.696 & 0.611 & 0.458 & 0.849 & 0.700 \\ 
        3 & + MixUp Training & \textbf{0.694} & \textbf{0.641} & 0.850 & \textbf{0.605} & \textbf{0.247} & 0.713 & \textbf{0.435} & 0.689 & \textbf{0.704} & \textbf{0.597} & 0.849 & \textbf{0.717} \\
        \bottomrule
    \end{tabular}
    \end{center}
    \caption{Effectiveness of reality-conforming scene composition and real-synthesis mixing up training strategy. (Evaluation dataset: nuScenes \emph{val} set, model: CenterPoint-Voxel, voxel size: [0.075,0.075,0.2])}
    \label{table: nusc_ap_different_step}
\end{table*}

The reality-conforming score $Re(\text{mAP})$ defined in Sec.~\ref{sec:reality_conforming_score} is deployed for comparing realisticness between GT-Aug and Real-Aug. Simultaneously, we ablate the contribution of each component in Real-Aug.
The detector, which possess a framework of CenterPoint-Voxel and a voxel size of [0.075,0.075,0.2], is trained on the vanilla nuScenes \emph{train} set.
The inference results on the augmented nuScenes \emph{val} set are compared and shown in Tab. \ref{table: nusc reality conforming score}. 
In contrast to GT-Aug, the $Re(\text{mAP})$ of Real-Aug increases from 0.744 to 0.933, which means our proposed reality-conforming scene composition approach and real-synthesis mixing up training strategy can effectively shrinkage the gap between the synthesized scenes and the real one. Each component, including the physically reasonable object position, heading, height and real scene-category relation, matters for realizing realistic scene synthesis for LiDAR augmentation in 3D object detection. The alignment of real scene-crowdedness relation finally regress to the raw point clouds without any synthesis-based augmentations.

\subsubsection{Effectiveness of Real-Aug on nuScenes \emph{val} set}
\label{sec:effectiveness of real-aug on nuscenes val set}

The effectiveness of Real-Aug, which contains a reality-conforming scene composition module and a real-synthesis mixing up training strategy, is validated on nuScenes \emph{val} set. 
Our proposed reality-conforming scene composition module boosts NDS from 0.666 to 0.678 and mAP from 0.595 to 0.611. 
Combining the real-synthesis mixing up training strategy, the performance of CenterPoint-Voxel can be further optimized and finally reaches 0.694 NDS and 0.641 mAP. 

The information of objects in nuScenes \emph{trainval} set are summarized in Appendix~\ref{sec:nuscenes object information} to help analyze the phenonmena shown in Tab.~\ref{table: nusc_ap_different_step}. 
In nuScenes dataset, car and pedestrian are two categories with most abundant data. Car exists in 97.73\% frames and pedestrian exists in 79.19\% frames. As a result, most detectors perform well on them. Although truck exists in 70.26\% frames, which ranks the 3rd place, the corresponding performance of detector is still worse than expectation. Trucks' unsatisfactory detection accuracy should be attributed to the rare low points density inside their bounding boxes. In each voxel with the size of [0.075,0.075,0.2], the average points number of truck is 0.044, which is mucher lower than that of barrier (0.550) and traffic cone (0.534). The above points-density-related issues are more serious in construction vehicle and trailer (with only 0.018 and 0.017 points per voxel). As a result, it is hard for detectors to distinguish them from background points. The abnormal mAP decline of motorcycle and bicycle when introducing GT-Aug also attract our attention, which may own to their complex morphology. 
As shown in Tab.~\ref{table: nusc_ap_different_step}, the effectivness of GT-Aug severely suffer from the dramatical mAP degradation of motorcycle and bicycle.

The proposed Real-Aug minimizes the misleading from non-existing LiDAR scan patterns introduced by GT-Aug. 
It excels at dealing with the complex-morphology and low-points-density issues, which is beneficial for unleashing the full power of detectors.
Replacing GT-Aug with Real-Aug achieves a boost of 6.3\%, 4.6\%, 11.7\%, 17.0\% AP for construction vehicle, trailer, motorcycle and bicycle respectively.

\subsubsection{Robustness of Real-Aug}
\label{sec:robustness of real-aug}

\setlength{\tabcolsep}{3pt}
\setlength{\doublerulesep}{2\arrayrulewidth}
\renewcommand{\arraystretch}{1.0}
\begin{table}[ht]
    \begin{center}
    \begin{tabular}{c|c|c|cc}
        \toprule
        Model & Method & voxel size & NDS & mAP \\ 
        \hline
        SECOND & GT-Aug & [0.1,0.1,0.2] & 0.620 & 0.505 \\
        SECOND & Real-Aug & [0.1,0.1,0.2] & 0.651 & 0.552\\
        \hline
        CenterPP & GT-Aug & [0.1,0.1,8.0] & 0.608 & 0.503 \\
        CenterPP & Real-Aug & [0.1,0.1,8.0] & 0.639 & 0.558 \\
        \hline
        CenterVoxel & GT-Aug & [0.075,0.075,0.2] & 0.666 & 0.595 \\
        CenterVoxel & Real-Aug & [0.075,0.075,0.2] & 0.694 & 0.641 \\
        \hline
        CenterVoxel & GT-Aug & [0.15,0.15,0.2] & 0.637 & 0.558 \\
        CenterVoxel & Real-Aug & [0.15,0.15,0.2] & 0.658 & 0.596 \\
        \bottomrule
    \end{tabular}
    \end{center}
    \caption{Model Robustness analysis. The results are evaluated on nuScenes \emph{val} set}
    \label{table: nusc different model}
\end{table}

\noindent\textbf{Robust to different detectors.}
The generality of Real-Aug is validated in multiple detectors with various voxel sizes. Evaluation results on nuScenes \emph{val} set are shown in Tab.~\ref{table: nusc different model}. 
In center-based models (including CenterPoint-Voxel and CenterPoint-Pillar), Real-Aug bring significant improvements (approximate 3\% NDS and 5\% mAP) over the baseline.
The optimized performance is also valid when decreasing the voxel resolution.
We test Real-Aug on SECOND-Multihead~\cite{CBGS}, which is a typical anchor-based detector, for further exploring its versatility.
The proposed realistic scene sysnthesis method for LiDAR augmentation also yields extra performance gain in anchor-based frameworks. 
It enhances SECOND-Multihead with a considerable increase of 3.1\% NDS and 4.7\% mAP.

\setlength{\tabcolsep}{5pt}
\setlength{\doublerulesep}{2\arrayrulewidth}
\renewcommand{\arraystretch}{1.0}
\begin{table}[ht]
    \begin{center}
    \begin{tabular}{c|c|ccc}
        \toprule
        Model & Method & Car & Pedestrian & Cyclist \\ 
        \hline
        SECOND & w/o GT-Aug & 77.8\% & 44.2\% & 56.5\% \\
        SECOND & GT-Aug & 81.4\% & 52.4\% & 65.3\% \\
        SECOND & Real-Aug & 81.7\% & 54.0\% & 68.2\% \\
        \hline
        PointPillar & w/o GT-Aug & 75.4\% & 42.1\% & 42.9\% \\
        PointPillar & GT-Aug & 77.9\% & 47.6\% & 63.2\% \\
        PointPillar & Real-Aug & 78.9\% & 51.5\% & 64.1\% \\
        \bottomrule
    \end{tabular}
    \end{center}
    \caption{Model Robustness analysis. The results are evaluated on KITTI \emph{val} set.}
    \label{table: kitti different model R40}
\end{table}

\noindent\textbf{Robust to different datasets.}
We test the adaptability of Real-Aug on KITTI dataset, in which objects and their distributions are highly divergent from that in nuScenes.
Thanks to the extensive expansion of training samples' diversity, detectors trained on KITTI can greatly benefit from GT-Aug and achieve better performance.
Even so, Real-Aug yields extra performance gain.
For the baseline of SECOND, GT-Aug increases $mAP_{3D}$ of moderate car, pedestrian and cyclist from 77.8\%, 44.2\%, 56.5\% to 81.4\%, 52.4\%, 65.3\%.
Replacing GT-Aug with Real-Aug, $mAP_{3D}$ can be further optimized to 81.7\%, 54.0\%, 68.2\%. 
Similar experimental phenomena are obtained when transforming detector from SECOND to PointPillar.
The $mAP_{3D}$ of all categories with various difficulties increases from 53.5\% to 62.9\% if GT-Aug is used. 
Real-Aug further enhances the performance of PointPillar to get a $mAP_{3D}$ of 64.8\%. 

\setlength{\tabcolsep}{3pt}
\setlength{\doublerulesep}{2\arrayrulewidth}
\renewcommand{\arraystretch}{1.0}
\begin{table}[ht]
    \begin{center}
    \begin{tabular}{c|c|c|cc}
        \toprule
        Method & sampled num & NDS & mAP \\ 
        \hline
        GT-Aug & ref*1 & 0.666 & 0.595 \\ 
        GT-Aug & ref*3 & 0.650 & 0.568 \\
        GT-Aug & fix=15 & 0.651 & 0.568 \\
        \hline
        Real-Aug & ref*1 & 0.694 & 0.641 \\
        Real-Aug & ref*3 & 0.693 & 0.639 \\
        Real-Aug & fix=15 & 0.690 & 0.634 \\
        \bottomrule
    \end{tabular}
    \end{center}
    \caption{Parameter robustness analysis. ref correspond to \{car:2, truck:3, construction vehicle.:7, bus:4, trailer:6, barrier:2, motorcycl':6, bicycle:6, pedestrian:2, traffic-cone:2\}. (Evaluation dataset: nuScenes \emph{val} set, model: CenterPoint-Voxel, voxel size: [0.075,0.075,0.2])}
    \label{table: nusc different hyper params}
\end{table}

\noindent\textbf{Robust to different hyper parameter choices.}
In GT-Aug, different magnitudes are used for sampling ground-truth objects with different categories. For each category, the magnitude means the number of objects that will be placed into a training point cloud frame. The reference design is proposed by Zhu et al.~\cite{CBGS} and widely used in the latter published works~\cite{CenterPoint, LargeKernel3D, LidarMultiNet, TransFusion}. We simply choose three magnitude design and compare their influence on the model trained with GT-Aug and Real-Aug. 
In Tab.~\ref{table: nusc different hyper params}, ref*1 denotes the same setting as previous optimized design~\cite{CBGS}, ref*3 means we multiply the magnitude for each category in~\cite{CBGS} by three times, fix=15 indicates we use the same setting 15 for all categories.
In GT-Aug, detectors' performance degrades greatly when magnitudes for different categories deviate from optimal solution. 
While Real-Aug is robust to different hyper parameter choices. 
According to our experience, only if there is enough inserted objects at the beginning stage of the model training and follow the real-synthesis mixing up training strategy given in Sec.~\ref{sec:real-synthesis mixing up}, 
the final results can almost reach the optimized level achieved by multiple attempts of hyper parameters for 10 catogories.

\subsubsection{Improvement of Data Utilization}
\label{sec:improvement of data utilization}

\setlength{\tabcolsep}{3pt}
\setlength{\doublerulesep}{2\arrayrulewidth}
\renewcommand{\arraystretch}{1.1}
\begin{table}[ht]
    \begin{center}
    \begin{tabular}{c|c|cc}
        \toprule
        Methods & $N_{train}/N_{total}$ & NDS & mAP \\ 
        \hline
        w/o GT-Aug & 1 & 0.659 & 0.594 \\ 
        GT-Aug & 1 & 0.666 & 0.595 \\ 
        Real-Aug & 1 & \textbf{0.694} & \textbf{0.641} \\
        Real-Aug & 1/2 & 0.681 & 0.622 \\
        Real-Aug & 1/4 & 0.667 & 0.600 \\
        Real-Aug & 1/8 & 0.641 & 0.555 \\
        \bottomrule
    \end{tabular}
    \end{center}
    \caption{Compare the performance of detectors trained with different proportions of data. (Evaluation dataset: nuScenes \emph{val} set, model: CenterPoint-Voxel, voxel size: [0.075,0.075,0.2]).}
    \label{table: nusc part train}
\end{table}

We compare the performance of detectors trained with different proportions of data and list them in Tab.~\ref{table: nusc part train}.
Real-Aug promotes the utilization of data to a great extent (with an approximate increase of 4 times).
The detector trained with 25\% data and Real-Aug performs comparably to the one trained with 100\% data and GT-Aug.

\subsubsection{Choice of different real-synthesis mixing up training strategy}
\label{sec:choice of training strategy}

\setlength{\tabcolsep}{3pt}
\setlength{\doublerulesep}{2\arrayrulewidth}
\renewcommand{\arraystretch}{1.0}
\begin{table}[ht]
    \begin{center}
    \begin{tabular}{c|c|c|cc}
        \toprule
        $\alpha$ start pct & $\beta$ div steps & $\beta$ div factor & NDS & mAP \\ 
        \hline
        - & - & - & 0.678 & 0.611 \\
        \hline
        0.75 & - & - & 0.688 & 0.629 \\
        \hline
        0.75 & [0.75,0.85] & 2 & \textbf{0.695} & \textbf{0.643} \\
        0.75 & [0.75,0.85] & 4 & 0.694 & 0.641 \\
        0.75 & [0.75,0.85] & 8 & 0.693 & 0.640 \\
        \hline
        0.80 & [0.80,0.90] & 2 & 0.693 & 0.639 \\
        0.80 & [0.80,0.90] & 4 & 0.695 & 0.640 \\
        0.80 & [0.80,0.90] & 8 & 0.695 & 0.642 \\
        \bottomrule
    \end{tabular}
    \end{center}
    \caption{Choice of different real-synthesis mixing-up training strategy. (Evaluation dataset: nuScenes \emph{val} set, model: CenterPoint-Voxel, voxel size: [0.075,0.075,0.2].) }
    \label{table: nusc different regress strategy}
\end{table}
 
we define two hyper parameters $\alpha$ and $\beta$ in Sec.~\ref{sec:real-synthesis mixing up} to align real scene-category and scene-crowdedness relation.
The divisity of synthesized scenes matters at the beginning of model training. 
The disturbance introduced by object insertion can be further modified by invoking our proposed real-synthesis mixing up training strategy.
The wake-up signal is defined by the start iteration percentage shown in the 1st and 2nd column of Tab.\ref{table: nusc different regress strategy}.
After receiving the signal, we linearly reduce $\alpha$ from 1.0 to 0.0 and use the step strategy to decrease $\beta$. 
\begin{figure*}[t]
\begin{center}
   \includegraphics[width=0.8\linewidth]{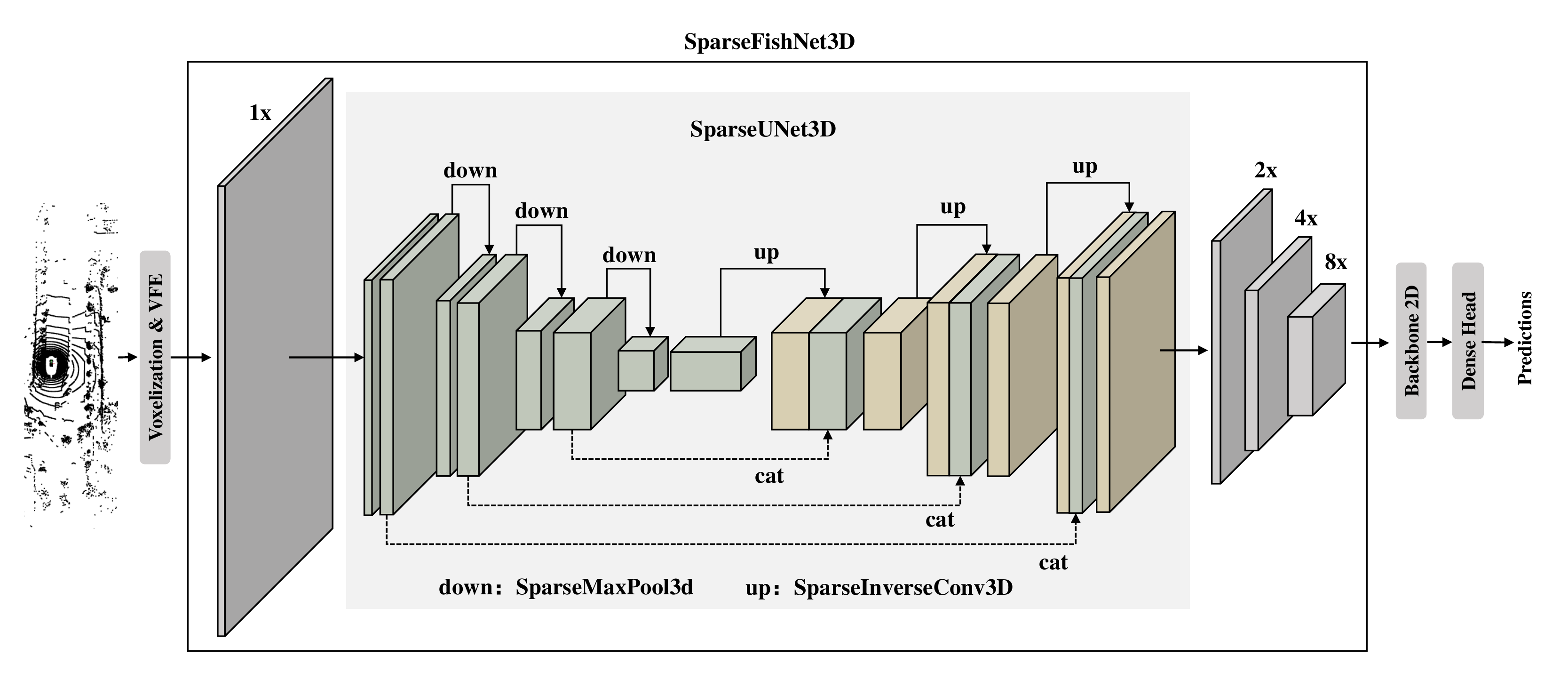}
\end{center}
   \caption{The framework of SparseFishNet3D. Based on the baseline backbone (SECOND), we apply a SparseUNet3D on the 2x downsample 3D feature map for stronger feature representation.}
\label{figure: fishnet3D}
\end{figure*}
As shown in Tab.~\ref{table: nusc different regress strategy}, real scene-category alignment leads to an increase of 1.0\% NDS and 1.8\% mAP. Combining the alignment to real scene-crowdedness relation, the highest NDS and mAP of CenterPoint-Voxel can reach 0.695 and 0.643.

\subsubsection{Backbone Optimization}
\label{sec:backbone optimization}

\setlength{\tabcolsep}{3pt}
\setlength{\doublerulesep}{2\arrayrulewidth}
\renewcommand{\arraystretch}{1.0}
\begin{table}[ht]
    \begin{center}
    \begin{tabular}{c|c|cc}
        \toprule
        Backbone & Real-Aug & NDS & mAP \\ 
        \hline
        SECOND(Baseline) & \ding{51} & 0.694 & 0.641 \\
        + SC-Conv & \ding{51} & 0.699 & 0.645 \\
        + IoU Pred & \ding{51} & 0.704 & 0.655 \\
        + SparseFishNet3D & \ding{51} & 0.710 & 0.661 \\
        \bottomrule
    \end{tabular}
    \end{center}
    \caption{Ablations on backbone optimization. The results are evaluated on nuScenes \emph{val} set}
    \label{table: nusc backbone optimization}
\end{table}

For enhancing the performance of detector, We use the self-calibrated convolution block (SC-Conv block) in 2D backbone and add an IoU prediction branch in the multi-task head as AFDetv2~\cite{AFDetV2}. The optimized detector achieves 1.6\% NDS and 2.0\% mAP increasement. We also apply a SparseUNet3D on the 2x downsample 3D feature map for stronger feature representation with larger receptive fields. Based on the above optimization methods, we achieve 0.710 NDS and 0.661 mAP on nuScenes \emph{val} set (without any test time augmentations).

\section{Conclusion}
A novel data-oriented approach, Real-Aug, is proposed for LiDAR-based 3D Object Detection. It consists of a reality-conforming scene composition module and a real-synthesis mixing up training strategy. We conduct extensive experiments to verify the effectiveness of Real-Aug and achieve a state-of-the-art 0.744 NDS and 0.702 mAP on nuScenes \emph{test} set. 

{\small
\bibliographystyle{ieee_fullname}
\bibliography{egbib}
}

\appendix

\clearpage
\section{Object Information}

In order to help analyze the difficulty of detection task in nuScenes and KITTI dataset, we summarize the basic information of objects with different categories. The results are shown in Tab.\ref{table: nusc object infos 2} and Tab.\ref{table: kitti object infos 2}. 

For a specific category, $\Bar{l}$, $\Bar{w}$, $\Bar{h}$ denotes the average length, width and height of objects. $\Bar{D}_{pts}$ denotes the average LiDAR points number inside each voxel of the object's 3D bounding box. It can be calculated by Eq.~\ref{eq:number of voxel}:

\begin{equation}
\label{eq:number of voxel}
    \Bar{\mathcal{D}}_{pts}=\frac{\Bar{N}_{pts}}{\Bar{N}_{voxel}} = \frac{\Bar{N}_{pts}}{(\Bar{l}/v_D) \times (\Bar{w}/v_W) \times (\Bar{h}/v_H)}
\end{equation}
where $\Bar{N}_{pts}$, $\Bar{N}_{voxel}$ denotes the average number of points, voxels inside object's 3D bounding box. $v_D$, $v_W$, $v_H$ are the voxel size defined in detectors. In Tab.~\ref{table: nusc object infos 2}, $v_D$, $v_W$, $v_H$ are set as 0.075, 0.075, 0.2. $\Bar{N}_{pts}$ is calculated according to the densified point cloud with 10 LiDAR sweeps. In Tab.~\ref{table: kitti object infos 2}, $v_D$, $v_W$, $v_H$ are set as 0.05, 0.05, 0.1. $\mathcal{R}_{frame}$ denotes the percentage of frames that contain the corresponding objects. The sum of $\mathcal{R}_{frame}$ for all classes is not equal to 1.0 because one frame may contain objects with different categories. $\mathcal{R}_{obj}$ denotes the percentage of objects throughout the whole \emph{object bank}.

\subsection{nuScenes}
\label{sec:nuscenes object information}

\setlength{\tabcolsep}{3pt}
\setlength{\doublerulesep}{2\arrayrulewidth}
\renewcommand{\arraystretch}{1.1}
\begin{table}[ht]
    \begin{center}
    \begin{tabular}{c|cccccc}
        \toprule
        \multirow{2}{*}{} & \multirow{2}{*}{$\Bar{l}$} & \multirow{2}{*}{$\Bar{w}$} & \multirow{2}{*}{$\Bar{h}$} & \multirow{2}{*}{$\Bar{\mathcal{D}}_{pts}$} & \multirow{2}{*}{$\mathcal{R}_{frame}$}& \multirow{2}{*}{$\mathcal{R}_{obj}$} \\ 
        \\
        \hline
        Car & 4.634 & 1.954 & 1.734 & 0.065 & 97.73\% & 42.43\% \\
        Truck & 6.992 & 2.517 & 2.870 & 0.044 & 70.26\% & 8.29\% \\
        C.V. & 6.454 & 2.857 & 3.216 & 0.018 & 23.33\% & 1.41\% \\
        Bus & 11.090 & 2.933 & 3.464 & 0.029 & 31.90\% & 1.60\% \\
        Trailer & 12.283 & 2.904 & 3.875 & 0.017 & 24.67\% & 2.40\% \\
        Barrier & 0.503 & 2.524 & 0.983 & 0.550 & 31.71\% & 13.65\% \\
        Mot. & 2.102 & 0.769 & 1.472 & 0.172 & 21.60\% & 1.16\% \\
        Byc. & 1.706 & 0.601 & 1.294 & 0.116 & 20.85\% & 1.07\% \\
        Ped. & 0.728 & 0.668 & 1.770 & 0.127 & 79.19\% & 20.11\% \\
        T.C. & 0.415 & 0.408 & 1.069 & 0.534 & 39.63\% & 7.88\% \\
        \bottomrule
    \end{tabular}
    \end{center}
    \caption{Information of objects in nuScenes \emph{trainval} set. }
    \label{table: nusc object infos 2}
\end{table}

According to the performance of CenterPoint-Voxel trained with GT-Aug, which is shown in the 2nd line of Tab.~\ref{table: nusc_ap_different_step}, we divide all the ten categories defined in nuScenes detection task into four groups: (1) car, pedestrian (with AP above 0.8); (2) bus, barrier, traffic cone (with AP ranging from 0.6 to 0.8); (3) truck, motorcycle, bicycle (with AP ranging from 0.4 to 0.6); (4) trailer, construction vehicle (with AP lower than 0.4).

Car and pedestrian are two categories with most abundant data in nuScenes dataset. Car exists in 97.73\% frames and pedestrian exists in 79.19\% frames. Their $\mathcal{R}_{obj}$ also ranks top two throughout the whole \emph{object band}. As a result, detectors perform well on car and pedestrian. Bus, barrier, traffic cone are three categories with clear and simple structural characteristics. The morphology consistency in different scenarios reduce the difficulty for detector to distinguish them from other objects and background points. Truck, motorcycle and bicycle, whose data is not as rich as pedestrian and meanwhile with more complex and volatile morphology than bus, barrier and traffic cone, are more difficult to be detected. For trailer and construction vehicle, the lowest points density in each voxel introduces much confusion for detectors to recognize them from background points. 

Real-Aug, which contains a a reality-conforming scene composition module to handle the details of the composition and a real-synthesis mixing up training strategy to gradually adapt the data distribution from synthetic data to real one, can greatly alleviate negative effects of existing synthesis-based augmentation methods. 
Detectors trained with Real-Aug present remarkable performance optimization, especially on motorcycle, bicycle, trailer and construction vehicle.

\subsection{KITTI}
\label{sec:kitti object information}

\setlength{\tabcolsep}{2pt}
\setlength{\doublerulesep}{2\arrayrulewidth}
\renewcommand{\arraystretch}{1.1}
\begin{table}[ht]
    \begin{center}
    \begin{tabular}{c|cccccc}
        \toprule
         & \multirow{2}{*}{$\Bar{l}$} & \multirow{2}{*}{$\Bar{w}$} & \multirow{2}{*}{$\Bar{h}$} & \multirow{2}{*}{$\Bar{\mathcal{D}}_{pts}$} & \multirow{2}{*}{$\mathcal{R}_{frame}$}& \multirow{2}{*}{$\mathcal{R}_{obj}$} \\ 
         \\
        \hline
        Car & 3.884 & 1.629 & 1.526 & 0.011 & 89.35\% & 82.46\% \\
        Ped. & 0.842 & 0.660 & 1.761 & 0.048 & 23.78\% & 12.87\% \\
        Cyc. & 1.764 & 0.597 & 1.737 & 0.023 & 15.25\% & 4.67\% \\
        \bottomrule
    \end{tabular}
    \end{center}
    \caption{Information of objects in KITTI \emph{trainval} set. }
    \label{table: kitti object infos 2}
\end{table}

According to the performance of SECOND, which is shown in Tab.~\ref{table: kitti different model R40}, we divide the three categories defined in KITTI detection task into two groups: (1) car; (2) pedestrian and cyclist. Comparing to car, pedestrian and cyclist are lack in data quantity and meanwhile possess high morphology complexity. As a result, the $mAP_{3D}$ of car are much higher than that of the other two categories. The effectiveness of Real-Aug is also validated in KITTI dataset. According to Tab.~\ref{table: kitti different model R40}, the $mAP_{3D}$ of moderate objects that is inferenced by SECOND can be further increased from 66.4\% to 68.0\% when replacing GT-Aug with Real-Aug.

\begin{figure*}[t]
\begin{center}
   \includegraphics[width=1.0\linewidth]{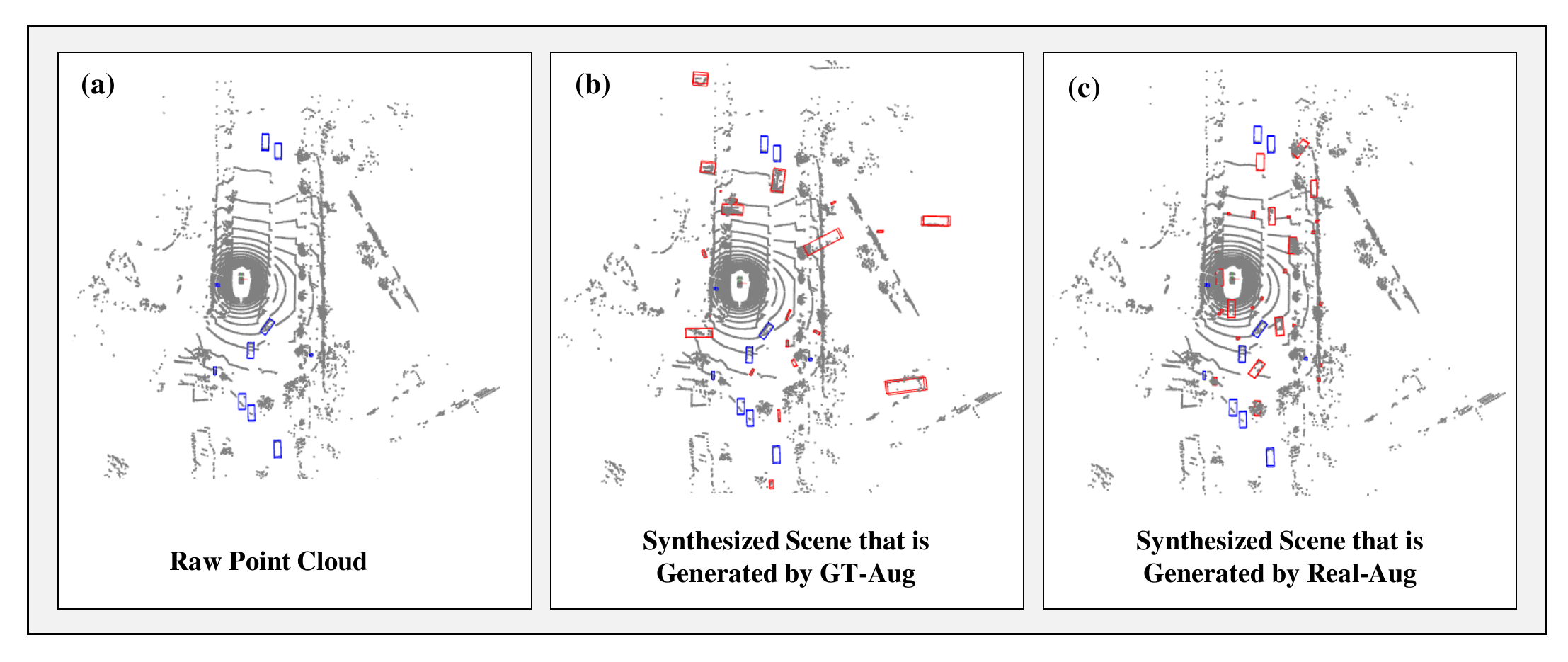}
\end{center}
   \caption{Visualization of synthesized scene generated by GT-Aug (b) and Real-Aug (c). (Boxes with blue lines: the original boxes in raw point clouds; Boxes with red lines: the inserted boxes.)}
\label{figure: aug cmp}
\end{figure*}

\begin{figure*}[t]
\begin{center}
   \includegraphics[width=0.8\linewidth]{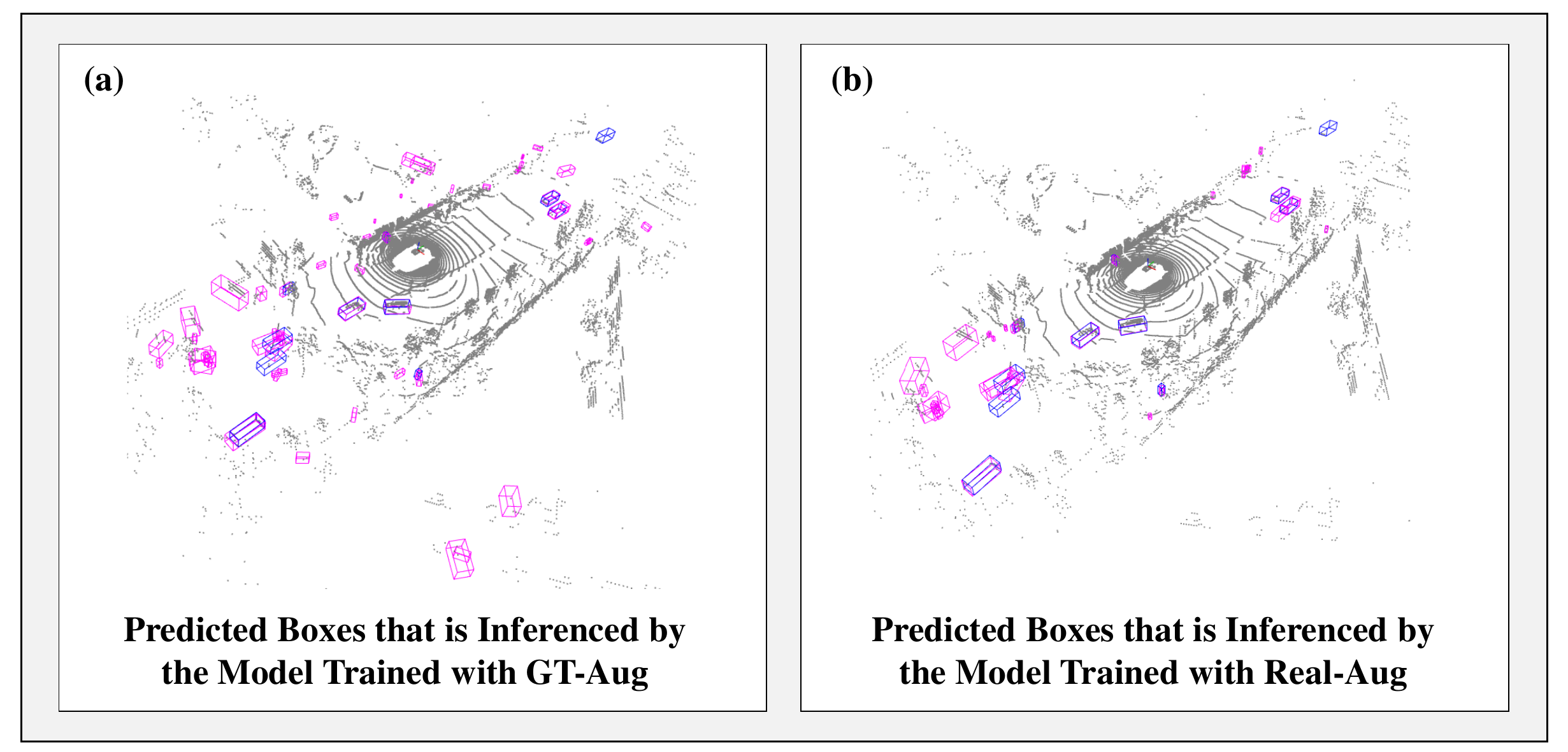}
\end{center}
   \caption{Visualization of predicted boxes that is inferenced by the model trained with GT-Aug (a) and Real-Aug (b). (Boxes with blue lines: ground-truth boxes; Boxes with pink lines: predicted boxes.)}
\label{figure: pred cmp}
\end{figure*}

\section{Test Time Augmentation}
\label{sec:test time augmentation}

\setlength{\tabcolsep}{3pt}
\setlength{\doublerulesep}{2\arrayrulewidth}
\renewcommand{\arraystretch}{1.1}
\begin{table}[ht]
    \begin{center}
    \begin{tabular}{c|c|c|cc}
        \toprule
        Model & double flip & rotation & NDS & mAP\\ 
        \hline
        CenterVoxel &  &  & 0.694 & 0.641 \\
        CenterVoxel & \ding{51} & \ding{51} & 0.715 & 0.667 \\
        SparseFishNet3D &  &  & 0.710 & 0.661 \\
        SparseFishNet3D & \ding{51} & \ding{51} & 0.731 & 0.689 \\
        \bottomrule
    \end{tabular}
    \end{center}
    \caption{Effects of TTA (double flip and rotation). (Evaluation dataset: nuScenes \emph{val} set, model:CenterPoint-Voxel, voxel size: [0.075,0.075,0.2]).}
    \label{table: nusc other tricks}
\end{table}

Throughout the inference process of CenterPoint-Voxel on nuScenes dataset, we use two test time augmentation (TTA), including double flip and point-cloud rotation along the yaw axis, to improve the detector's final detection performance. The yaw angles are set the same as that in \cite{CenterPoint}, which are $[0^\circ,\pm 6.25^\circ,\pm 12.5^\circ,\pm 25^\circ]$. The inference results of CenterPoint-Voxel on nuScenes \emph{val} set are shown in Tab.~\ref{table: nusc other tricks}. 
With TTA, the NDS and mAP of the baseline model (CenterPoint-Voxel) raise from 0.694 and 0.641 to 0.715 and 0.667 respectively. The SparseFishNet3D which is decribed in Sec.~\ref{sec:backbone optimization} further enhances the detection performance with a considerable 0.731 NDS and 0.689 mAP.

\setlength{\tabcolsep}{5pt}
\setlength{\doublerulesep}{2\arrayrulewidth}
\renewcommand{\arraystretch}{1.0}
\begin{table}[ht]
    \begin{center}
    \begin{tabular}{c|c|ccc}
        \toprule
        Model & y-flip & Car & Pedestrian & Cyclist \\ 
        \hline
        SECOND &  & 81.7\% & 54.0\% & 68.2\% \\
        SECOND & \ding{51} & 81.8\% & 57.3\% & 68.6\% \\ 
        \bottomrule
    \end{tabular}
    \end{center}
    \caption{Effects of TTA (y-flip). (Evaluation dataset: KITTI \emph{val} set, model:SECOND, voxel size: [0.05,0.05,0.1]).}
    \label{table: kitti other tricks R40}
\end{table}

In KITTI dataset, objects outside the front view are not annotated. Throughout the inference process of SECOND, y-flip test time augmentation is applied. The evaluated results of SECOND on KITTI \emph{val} set are shown in Tab.~\ref{table: kitti other tricks R40}. There is an increase of 0.1\%, 3.3\%, 0.4\% $AP_{3D}$ for moderate car, pedestrian and cyclist. 

\section{Visualization}

The augmented point clouds generated by GT-Aug and Real-Aug are visualized in Fig.~\ref{figure: aug cmp}. GT-Aug introduces many non-existing LiDAR scans patterns into the point clouds. The inserted objects, which locate at physically unreasonable place and move towards inappropriate direction, will hinder detectors from learning effective features. In this paper, a reality-conforming scene composition module is proposed to deal with the above mentioned problems. It handles the details of synthesis operation and maintains the authenticity of the composite scene as much as possible. The real-synthesis mixing up training strategy can further alleviate the negative influence introduced by synthesis-based LiDAR augmentation. The predicted boxes that is inferenced by the model trained with GT-Aug and Real-Aug are visualized in Fig.~\ref{figure: pred cmp}. The predicted boxes with scores lower than 0.1 are filtered out. It is clear that replacing GT-Aug with Real-Aug can effectively reduce false positives.

\section{Limitations and Future Work}

In addition to applying our Real-Aug on nuScenes and KITTI dataset, we will also extend our methods on Waymo dataset in the future.

\end{document}